\documentclass[10pt,twocolumn,letterpaper]{article}
\usepackage{times}
\usepackage{cvpr}              

\definecolor{cvprblue}{rgb}{0.21,0.49,0.74}
\usepackage[pagebackref,breaklinks,colorlinks,allcolors=cvprblue]{hyperref}

\def\paperID{ABAW-67} 
\def\confName{CVPR}
\def\confYear{2025}

\title{MAVEN: Multi-modal Attention for Valence-Arousal Emotion Network}

\author{
Vrushank Ahire\textsuperscript{$1$}\\
{\tt\small 2022csb1002@iitrpr.ac.in}
\and
Kunal Shah\textsuperscript{$1$}\\
{\tt\small 2023aim1003@iitrpr.ac.in}
\and
Mudasir Nazir Khan\textsuperscript{$1$}\\
{\tt\small 2024aim1006@iitrpr.ac.in}
\and
Nikhil Pakhale\textsuperscript{$1$}\\
{\tt\small 2024dss1013@iitrpr.ac.in}
\and
Lownish Rai Sookha\textsuperscript{$1$}\\
{\tt\small lownish.23csz0010@iitrpr.ac.in}
\and
M. A. Ganaie\textsuperscript{$1$}\\
{\tt\small mudasir@iitrpr.ac.in}
\and
Abhinav Dhall\textsuperscript{$2$}\\
{\tt\small abhinav.dhall@monash.edu}
\vspace{0.3cm}
\\
\textsuperscript{$1$}Department of Computer Science and Engineering, Indian Institute of Technology Ropar, Punjab, India\\
\textsuperscript{$2$}Department of Data Science and AI, Monash University, Melbourne, Australia
}

\begin{document}
\maketitle
\begin{abstract}
Dynamic emotion recognition in the wild remains challenging due to the transient nature of emotional expressions and temporal misalignment of multi-modal cues. Traditional approaches predict valence and arousal and often overlook the inherent correlation between these two dimensions. The proposed Multi-modal Attention for Valence-Arousal Emotion Network (MAVEN) integrates visual, audio, and textual modalities through a bi-directional cross-modal attention mechanism. MAVEN uses modality-specific encoders to extract features from synchronized video frames, audio segments, and transcripts, predicting emotions in polar coordinates following Russell’s circumplex model. The evaluation of the Aff-Wild2 dataset using MAVEN achieved a concordance correlation coefficient (CCC) of 0.3061, surpassing the ResNet-50 baseline model with a CCC of 0.22. The multistage architecture captures the subtle and transient nature of emotional expressions in conversational videos and improves emotion recognition in real-world situations.
\end{abstract}  
\section{Introduction}
\label{sec:intro}

 Emotion recognition is a critical challenge in affective computing with applications spanning human-computer interaction, healthcare, education, and entertainment. While traditional approaches have focused on categorical emotion classification (e.g., happiness, sadness, anger), dimensional models representing emotions along continuous valence (pleasure-displeasure) and arousal (activation-deactivation) scales have gained prominence for their ability to represent subtle emotional nuances~\cite{verma2017affect, kollias2019deep}. These dimensional models support the idea that emotions exist on a continuous spectrum, enriching the understanding of the complexity of human emotional experiences. 

The detection and analysis of human emotions in natural settings pose significant challenges due to variations in expressions, individual differences, cultural influences, and the often subtle nature of emotional cues~\cite{kollias2021affect,zafeiriou2017aff}. This complexity requires a multi-modal approach, as people naturally express and perceive emotions through various channels, including facial expressions, voice intonation, language, and gestures~\cite{cimtay2020cross, zhang2024deep}. Integrating complementary information from each modality enhances the robustness and accuracy of emotion recognition systems. While facial expressions might reveal visible emotional cues, speech prosody can uncover subtle emotional undertones, and linguistic content can provide contextual information essential for accurate interpretation~\cite{khan2024mser, meng2022multi}. 

The field has witnessed significant advancements in emotion recognition through deep learning approaches in recent years. Initial research emphasizes unimodal systems in facial expression analysis using efficient deep learning models like EfficientNet and extracting frame-level features through EmotiEffNet~\cite{savchenko2022frame,savchenko2023emotieffnets}. Approaches such as EmoFAN-VR achieved notable success in controlled environments but often struggled with in-the-wild scenarios characterized by varying illumination, occlusions, and pose variations~\cite{kollias2019deep,gotsman2021valence}. Researchers explored other modalities separately, creating specialized models for emotion recognition in speech using spectral features and recurrent architectures such as Vesper~\cite{chen2024vesper} and text-based systems using natural language processing techniques. 

The shift towards multi-modal systems reflects how humans perceive and express emotions through various channels simultaneously~\cite{zhang2024deep}. Initial multi-modal approaches employed simple fusion strategies, such as feature concatenation or decision-level integration, which failed to capture the inter-dependencies between modalities~\cite{cimtay2020cross}.  Advanced techniques were developed, including tensor-based fusion, specifically Tucker Tensor Regression and Tensor Regression Networks~\cite{mitenkova2019valence}, bilinear pooling, and various ensemble techniques like EVAEF~\cite{liu2023evaef} . While these approaches have enhanced performance, they often treat different modalities as independent sources of information.  To address this, Meng et al.~\cite{meng2022valence} uses temporal encoders, specifically transformer-based and Long Short-Term Memory (LSTM)-based networks, to better understand how emotions evolve over time in a video.

Attention mechanisms have emerged as a promising approach to address these limitations by enabling models to focus on relevant features across modalities. Early attention-based methods primarily implemented self-attention within individual modalities or simple cross-modal attention between pairs of modalities. Praveen et al.~\cite{praveen2022joint} implemented a joint cross-attention fusion model, and Zhang et al.~\cite{zhang2023abaw5} introduced TEMMA, a multi-head attention module. These approaches demonstrated improved performance but created limited pathways for information exchange, failing to utilize the complementary nature of multi-modal data completely. Most existing works have concentrated on categorical emotion recognition~\cite{kollias2019expression}, with comparatively less exploration of attention mechanisms for continuous valence-arousal prediction.

The Affective Behavior Analysis in-the-wild (ABAW) competitions have significantly advanced research in this domain by providing standardized benchmarks and challenging in-the-wild datasets~\cite{kollias2022abaw, kollias2023abaw, kollias20246th, kollias20247th}. Despite advancements, several significant challenges still remain unaddressed. First, existing fusion approaches often struggle to combine information across modalities while maintaining temporal coherence effectively~\cite{dresvyanskiy2024multi}. Second, attention mechanisms have not been fully exploited to capture the complex inter-relationships between different emotional cues~\cite{yu2024improving}. Third, the direct regression of valence-arousal values in Cartesian coordinates may not optimally align with psychological models of emotion~\cite{verma2017affect}. 
\begin{figure}
    \centering
    \includegraphics[width=0.4\textwidth]{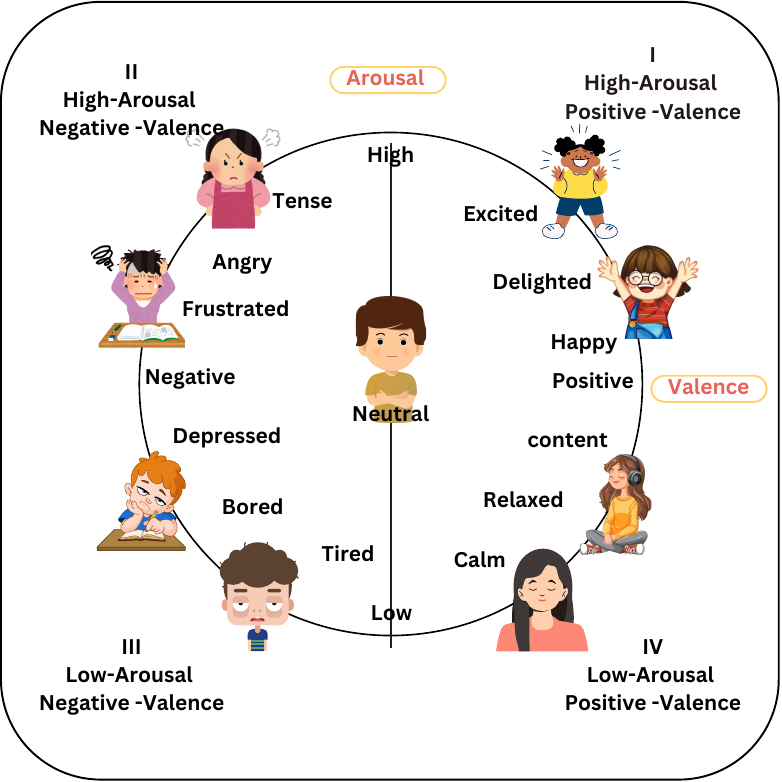}
    \caption{Valence-Arousal Emotion Circumplex}
    \label{fig:VA_Image}
\end{figure}
We propose a novel Multi-modal Attention for Valence-Arousal Emotion Network (MAVEN) with several key innovations to address these limitations. The key contributions are:
\begin{itemize}
\item We employ state-of-the-art (SOTA) modality-specific encoders: Swin Transformer for visual data, HuBERT for audio, and RoBERTa for text, to extract robust feature representations from each data stream.
\item  The proposed model employs a cross-modal attention mechanism that uses six distinct attention pathways and enables interactions between all modality pairs through weighted attention from other modalities. This design allows each modality to both inform and be informed by others, creating a rich information exchange network.
\item   This is followed by self-attention within its modality-specific encoder, utilizing a multi-headed attention module akin to that used in the Bidirectional Encoder Representations from Transformers (BEiT) model.
\item Additionally, we exploit polar coordinate form, representing emotions as angle ($\theta$) and intensity ($I$) rather than directly predicting valence-arousal values. This representation aligns naturally with psychological models of the emotion circumplex~\cite{russell1980circumplex}.
\item Extensive experiments on the Aff-Wild2 dataset~\cite{kollias2021affect} demonstrate that the proposed approach significantly outperforms existing methods as measured by the Concordance Correlation Coefficient (CCC)~\cite{kollias2021analysing}. 
\end{itemize}

The results validate the effectiveness of our bidirectional multi-modal attention architecture and polar coordinate prediction framework for capturing complex emotional expressions in conversational videos.

This paper is organized as follows: Section 2 reviews related work on emotion recognition, multi-modal fusion, and attention mechanisms. Section 3 presents our proposed methodology, which includes the multi-modal attention architecture and the polar coordinate prediction framework. Section 4 details the experimental setup, results based on the Aff-Wild2 dataset, and the ablation studies and analyzes the contributions of individual components. Finally, Section 5 concludes the paper and discusses future directions.

\section{Related Work}
\label{sec:related_works}
This section reviews the evolution of emotion recognition approaches, starting with traditional unimodal methods, progressing to multi-modal fusion techniques, and examining the role of attention mechanisms. It also explores the shift from categorical to dimensional models of emotion representation.

\subsection{Unimodal Emotion Recognition}
Early research in emotion recognition often focused on single modalities such as visual cues, audio signals, or textual content.

\subsubsection{Visual-Based Approaches}
Visual emotion recognition primarily focuses on analyzing facial expressions and body gestures. Early methods relied on handcrafted features to capture spatio-temporal information, but these often required significant prior knowledge and demonstrated a disconnect between the extracted features and actual emotional expressions. Recently, deep learning models, particularly Convolutional Neural Networks (CNNs) and Recurrent Neural Networks (RNNs), have become the preferred approach~\cite{zhang2023abaw5,khan2024mser,zhou2025emotion}. For facial expressions, models like SISTCM (Super Image-based Spatio-Temporal Convolutional Model)~\cite{song2018spatiotemporal} and two-stream LSTM models have been developed to capture local spatio-temporal features and global temporal cues related to emotional changes, with SISTCM utilizing 2D convolution for efficiency. The advancements in body gesture recognition have emerged through representation methods based on body joint movements, including the Attention-based Channel-wise Convolutional Model (ACCM), which employs channel-wise convolution and attention mechanisms to learn key joint features. Beyond facial expressions and body gestures, other visual cues like eye movement and body posture also play a significant role in emotion recognition~\cite{fan2024icaps}.

\subsubsection{Audio-Based Approaches} 
 Audio-based emotion recognition, also known as Speech Emotion Recognition (SER), involves analyzing speech signals to identify emotional states. Traditional methods focused on extracting low-level features such as Mel-frequency cepstral coefficients (MFCCs), pitch, and energy~\cite{chen2024vesper}. Statistical analysis was then frequently applied to these handcrafted features. However, with the advancement of deep learning, techniques such as Deep Neural Networks (DNNs), CNNs using spectrograms or MFCCs as inputs, and RNNs like LSTM networks and Gated Recurrent Units (GRUs) have demonstrated significant improvements in SER performance. Additionally, attention mechanisms have gained popularity in this field, allowing models to focus on the most important parts of the speech signal.

\subsubsection{Text-Based Approaches}
Text-based emotion recognition, often referred to as sentiment analysis, examines written text to identify the emotions or sentiments expressed within it. Early methods primarily relied on lexicon-based approaches and traditional machine learning classifiers, such as Support Vector Machines, which were trained using features like bag-of-words or Term Frequency - Inverse Document Frequency (TF-IDF). Deep learning techniques have revolutionized text-based emotion recognition. CNNs can extract local patterns, while RNNs, LSTMs, and GRUs- effectively capture sequential dependencies in text. Pre-trained word embeddings like Word2Vec and GloVe are used to represent words semantically. Transformer networks have also achieved SOTA results in text-based emotion recognition by effectively modeling long-range dependencies and contextual -modal approaches, which can be affected by noise, occlusions, or the masking of emotions~\cite{kollias2019deep}, researchers have increasingly turned their attention to multi-modal emotion recognition (MER)~\cite{cimtay2020cross}. MER utilizes information from various modalities, including audio, visual (such as facial expressions and body gestures), and text, to achieve more accurate emotion recognition. The fundamental principle behind this approach is that different modalities can provide complementary insights into human emotions.

Various multi-modal fusion techniques have been explored:
\begin{itemize}
    \item Early Fusion (Feature Fusion): This approach involves concatenating features extracted from different modalities at an early stage and then feeding the combined feature vector into a classifier. For example, audio and visual features might be extracted using CNNs and then combined before being processed by an RNN.
    \item Late Fusion (Decision Fusion)\textbf{:} In late fusion, separate classifiers are trained for each modality, and their individual predictions (e.g., probability scores) are then combined using methods like averaging, weighted averaging, or voting to make a final emotion prediction.
    \item Hybrid Approaches: These methods combine aspects of early and late fusion. For instance, features from different modalities might be processed separately by subnetworks, and the resulting intermediate representations are then fused before the final classification layer.
\end{itemize}

Deep learning models have been instrumental in developing advanced multi-modal fusion strategies. Techniques like concatenation, element-wise addition or multiplication, and more sophisticated methods using attention mechanisms or bilinear pooling are commonly employed to integrate information from multiple modalities.

\begin{figure}
    \centering
    \includegraphics[width=0.38\textwidth]{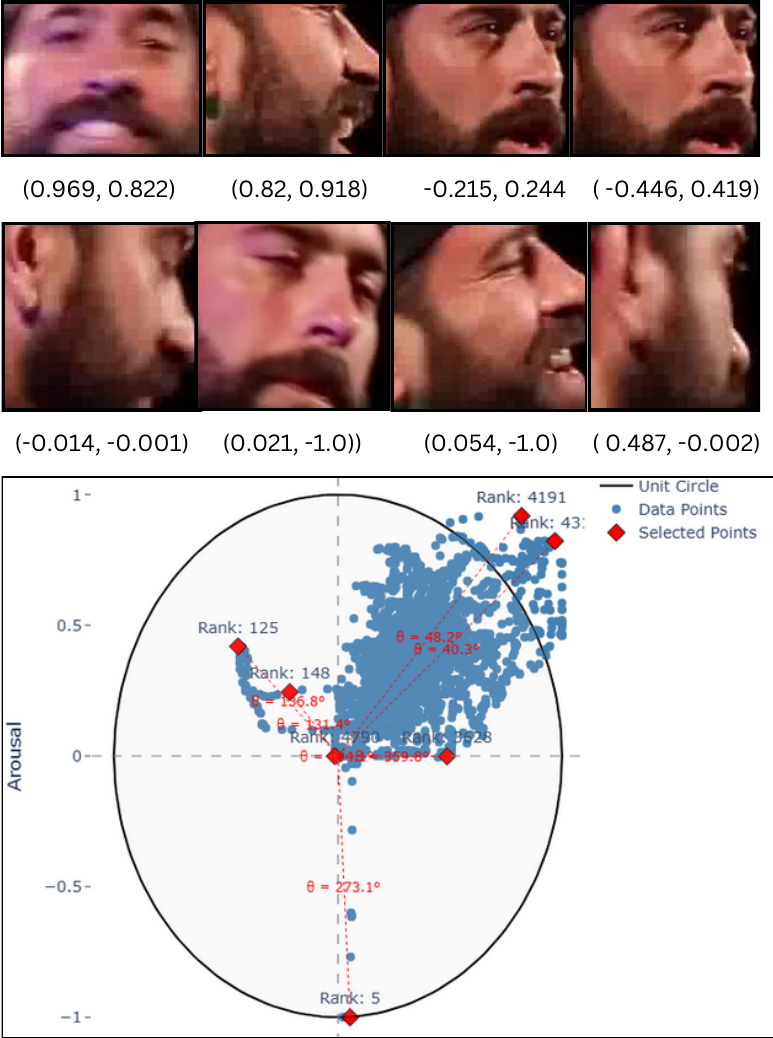}
    \caption{Valence-Arousal Distribution for a video sample from the Aff-Wild2 dataset}
    \label{fig:face_images}
\end{figure}

\subsection{Attention Mechanisms for Emotion Recognition}
Attention mechanisms have become a crucial component in modern emotion recognition systems, enabling models to selectively focus on the most relevant parts of the input data across different modalities and time steps~\cite{liu2023evaef,kollias20246th}. In visual emotion recognition, attention can help the model focus on emotion-relevant facial regions or specific body joints. In audio processing, attention mechanisms can weigh different parts of the speech signal based on their importance for emotion recognition. For text analysis, attention allows the model to identify the words that are most indicative of the expressed emotion.

In multi-modal emotion recognition, attention mechanisms play a vital role in learning the inter-modal relationships and determining the contribution of each modality to the final prediction. Cross-attention mechanisms allow the model to attend to relevant information in one modality based on the content of another modality, facilitating efficient information fusion. Self-attention mechanisms enable the model to capture intra-modal dependencies by attending to different parts of the same modality. Multi-head attention, as used in Transformer networks~\cite{dresvyanskiy2024multi}, allows the model to attend to different aspects of the input simultaneously.

\subsection{Dimensional Models of Emotion}

Early research in emotion recognition primarily focused on identifying a specific set of basic emotions, such as anger, disgust, fear, happiness, sadness, surprise, and sometimes neutrality~\cite{zhang2023abaw5}. However, researchers have shifted towards dimensional models of emotion ~\cite{savran2013automatic,mitenkova2019valence}. These models depict emotions as points within a continuous space characterized by dimensions like valence (pleasantness) and arousal (intensity)~\cite{zafeiriou2017aff}. Additionally, some models incorporate a third dimension, which is dominance (the level of control).

The Valence-Arousal (VA) space is the most commonly used dimensional representation of emotions. In this model, valence ranges from negative to positive, while arousal varies from passive to active. This two-dimensional framework allows for a circumplex representation of emotions, where different specific emotions can be mapped to distinct regions within the space. The Affective Behavior Analysis in-the-wild (ABAW) competitions have played a significant role in advancing valence-arousal estimation in real-world scenarios by providing large-scale annotated datasets, such as Aff-Wild and Aff-Wild2~\cite{kollias2022abaw,kollias2023abaw,kollias20246th,kollias20247th,kollias2025advancements}.

Building on previous advancements, we propose MAVEN, a novel multi-modal attention framework for valence-arousal estimation. The next section describes our proposed model.

\section{Proposed Model: MAVEN} 
\label{sec:proposed_model}

\begin{figure*}[!t]
    \centering
    \includegraphics[width=\textwidth]{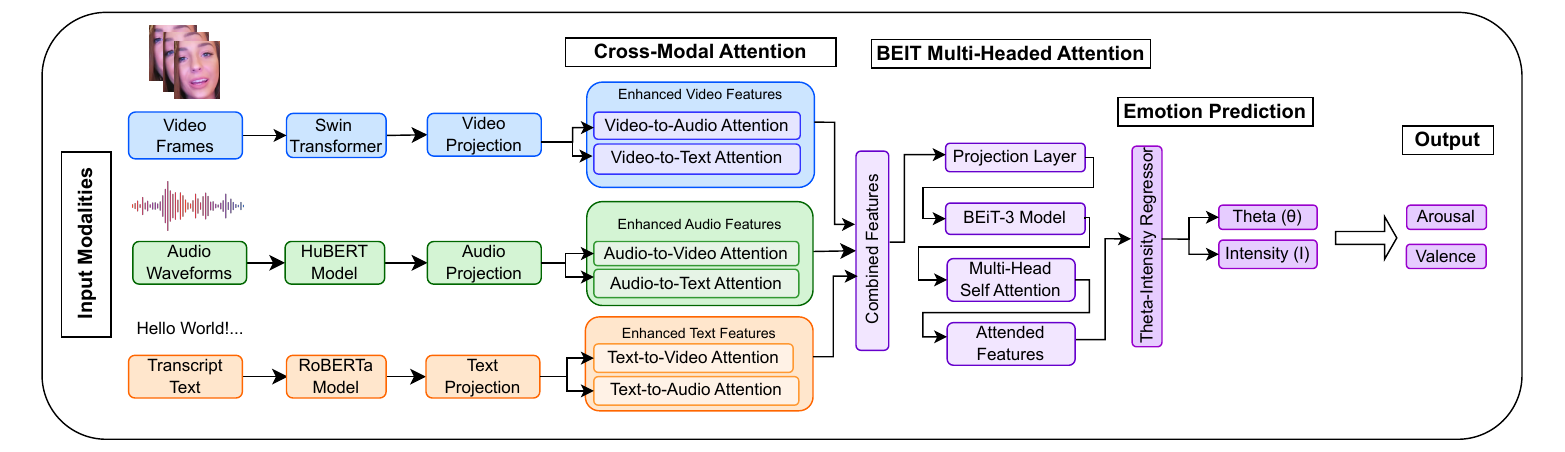}
    \caption{Multi-modal Attention for Valence-Arousal Emotion Network (MAVEN) Architecture. The model processes visual, audio, and text inputs through specialized feature extractors, fuses information via cross-modal attention pathways, refines representations with bidirectional self-attention, and predicts emotions using a polar coordinate system.}
    \label{fig:maven_architecture}
\end{figure*}

This section details MAVEN (Multi-modal Attention for Valence-arousal Emotion Network), our novel approach to continuous emotion recognition in conversational videos. Figure~\ref{fig:maven_architecture} illustrates the overall architecture of our proposed model.

\subsection{Overview}
MAVEN integrates information from three modalities: visual, audio, and text, employing an attention mechanism that enhances the exchange of information between these modalities. The system consists of five key components:
\begin{itemize}
    \item Modality-specific feature extractors for visual, audio, and textual data
    \item A comprehensive cross-modal attention mechanism with bidirectional information flow
    \item A bidirectional multi-headed self-attention module for each modality
    \item A BEiT-based encoder refinement layer
    \item A polar coordinate-based emotion prediction framework
\end{itemize}

\subsection*{Modality-Specific Feature Extraction}

\subsection{Visual Features}
We employ the Swin Transformer~\cite{liu2021swin} to extract visual features from facial regions in each video frame. The Swin Transformer combines local attention with shifted windows, enabling efficient modeling of hierarchical visual features while maintaining linear computational complexity relative to image size.

\subsubsection{Patch Embedding}
Given a video sequence $\{V_1, V_2, ..., V_T\}$ with $T$ frames, each frame $V_i \in \mathbb{R}^{H \times W \times 3}$ is split into non-overlapping patches of size $P \times P$:

\begin{equation}
X_p = \text{PatchPartition}(V_i) \in \mathbb{R}^{\frac{H W}{P^2} \times (P^2 \cdot 3)}
\end{equation}

These patches are then projected into a $d_v$-dimensional embedding space:

\begin{equation}
Z_0 = X_p W_p + b_p, \quad Z_0 \in \mathbb{R}^{\frac{H W}{P^2} \times d_v}
\end{equation}

where $W_p \in \mathbb{R}^{(P^2 \cdot 3) \times d_v}$ and $b_p \in \mathbb{R}^{d_v}$ are learnable parameters.

\subsubsection{Shifted Window Multi-Head Self-Attention (SW-MSA)} 
To model long-range dependencies efficiently, attention is applied within local windows that shift across layers to capture cross-window interactions:

\begin{equation}
\text{Attention}(Q, K, V) = \text{softmax} \left( \frac{QK^T}{\sqrt{d_k}} \right) V
\end{equation}

where $Q, K, V = Z_{l-1} W_Q, Z_{l-1} W_K, Z_{l-1} W_V$, with $d_k = d_v/h$ and $h$ as the number of attention heads.

\subsubsection{Visual Feature Output} 
After processing through $L$ transformer layers with hierarchical merging, we obtain the final visual feature sequence:

\begin{equation}
F_V = \text{SwinT}(V_1, V_2, ..., V_T) \in \mathbb{R}^{T \times d_v}
\end{equation}

where $F_V$ represents the sequence of visual features with dimension $d_v = 1024$.

\subsection{Audio Features} 
For audio processing, we utilize HuBERT~\cite{hsu2021hubert}, a self-supervised speech representation learning model with superior performance in capturing acoustic characteristics relevant to emotion recognition.

\subsubsection{Feature Extraction}
Given the audio signal \( A(t) \) corresponding to the video frames, we first compute log-mel spectrogram features:
\begin{equation}
X_A = \text{MelFilterBank}(\text{STFT}(A(t))) \in \mathbb{R}^{T' \times F}, \label{eq:log_mel}
\end{equation}
where \( T' \) is the number of time frames in the spectrogram and \( F \) is the number of mel frequency bands.

where $T'$ is the number of time frames in the spectrogram and $F$ is the number of mel frequency bands.

\subsubsection{Audio Encoder} 
A CNN maps $X_A$ to initial hidden representations:

\begin{equation}
Z_A = E_{a}(X_A) \in \mathbb{R}^{T'' \times d_a}
\end{equation}

where $T''$ is the reduced temporal dimension and $d_a$ is the feature dimension.

\subsubsection{Audio Feature Output} 
After processing through the HuBERT model, we obtain the final audio feature sequence:

\begin{equation}
F_A = \text{HuBERT}(A_1, A_2, ..., A_T) \in \mathbb{R}^{T \times d_a}
\end{equation}

where $F_A$ represents the sequence of audio features with dimension $d_a = 768$.

\subsection{Text Features}
We employ RoBERTa~\cite{liu2019roberta}, a robust variant of BERT, to extract semantic features from transcribed speech, capturing emotional markers from linguistic content.

\subsubsection{Token Embedding}
For a sequence of tokens $\{t_1, t_2, ..., t_S\}$ derived from the transcribed speech, we compute embeddings as:

\begin{equation}
E_t = E_{\text{token}}(t_i) + E_{\text{pos}}(i), \quad E_t \in \mathbb{R}^{S \times d_t}
\end{equation}

where $E_{\text{token}}$ is the token embedding function, $E_{\text{pos}}$ is the positional embedding function, and $d_t$ is the embedding dimension.

\subsubsection{Multi-Head Self-Attention}
The token embeddings are processed through multiple layers of self-attention:
\begin{align}
\text{MHA}(Q, K, V) &= \text{Concat}(\text{head}_1, \dots, \text{head}_h)W_O, \label{eq:mha} \\
\text{head}_i &= \text{Attention}(Q W_i^Q, K W_i^K, V W_i^V), \label{eq:head_i}
\end{align}
where each attention head computes the above for \( i = 1, \dots, h \).

\subsubsection{Text Feature Output} 
The processed textual features are obtained as:

\begin{equation}
F_T = \text{RoBERTa}(t_1, t_2, ..., t_S) \in \mathbb{R}^{S \times d_t}
\end{equation}

where $F_T$ represents the sequence of text features with dimension $d_t = 768$.

To align text features with visual and audio features that have temporal dimension $T$, we apply temporal interpolation:

\begin{equation}
F_T' = \text{TemporalInterpolate}(F_T) \in \mathbb{R}^{T \times d_t}
\end{equation}

\subsection{Cross-Modal Attention}

The core contribution of our approach is the bidirectional cross-modal attention mechanism, which enables information exchange between all pairs of modalities. This mechanism generates six distinct attention pathways: (visual $\rightarrow$ audio, visual $\rightarrow$ text, audio $\rightarrow$ visual, audio $\rightarrow$ text, text $\rightarrow$ visual, text $\rightarrow$ audio). This allows each modality to both provide information to and receive information from the others.

For each modality pair $(m, n)$ where $m, n \in \{\text{visual}, \text{audio}, \text{text}\}$ and $m \neq n$, we compute cross-modal attention as follows:

\begin{equation}
\text{Cross-Attention}(Q_m, K_n, V_n) = \text{softmax}\left( \frac{Q_m K_n^T}{\sqrt{d_k}} \right) V_n
\end{equation}
where:
\begin{itemize}
    \item $Q_m = F_m W_Q^m$ is the query matrix derived from modality $m$
    \item $K_n = F_n W_K^n$ is the key matrix derived from modality $n$
    \item $V_n = F_n W_V^n$ is the value matrix derived from modality $n$
    \item $W_Q^m \in \mathbb{R}^{d_m \times d_k}$, $W_K^n \in \mathbb{R}^{d_n \times d_k}$, and $W_V^n \in \mathbb{R}^{d_n \times d_v}$ are learnable parameter matrices
    \item $d_k$ is the dimension of the keys and queries
\end{itemize}

For example, the attention from audio to visual features is computed as:  
\begin{equation}
A_{A \rightarrow V} = \text{Cross-Attention}(W_Q^A F_A, W_K^V F_V, W_V^V F_V)
\end{equation}

Similarly, we compute attention for the other five directional pathways, where \( A_{V \rightarrow A} \), \( A_{V \rightarrow T} \), \( A_{A \rightarrow T} \), \( A_{T \rightarrow V} \), and \( A_{T \rightarrow A} \) are obtained via cross-attention mechanisms using modality-specific query, key, and value projections of \( F_V \), \( F_A \), and \( F_T' \), with dimensions \( A_{A \rightarrow V}, A_{T \rightarrow V} \in \mathbb{R}^{T \times d_v} \), \( A_{V \rightarrow A}, A_{T \rightarrow A} \in \mathbb{R}^{T \times d_a} \), and \( A_{V \rightarrow T}, A_{A \rightarrow T} \in \mathbb{R}^{T \times d_t} \).









\subsection{BEiT-based Encoder Refinement}
Following cross-modal attention, we apply a two-step refinement process: (1) Bidirectional Self-Attention Refinement, where intra-modal dependencies are strengthened, and (2) BEiT-based Encoder Refinement, which further contextualizes and integrates multi-modal features. To enhance intra-modal relationships, we apply self-attention within each modality:
\begin{align}
F_V^{\text{refined}} &= \text{SelfAtt}(F_V^{\text{enhanced}}) \in \mathbb{R}^{T \times d_v}, \label{eq:F_V_refined} \\
F_A^{\text{refined}} &= \text{SelfAtt}(F_A^{\text{enhanced}}) \in \mathbb{R}^{T \times d_a}, \label{eq:F_A_refined} \\
F_T^{\text{refined}} &= \text{SelfAtt}(F_T^{\text{enhanced}}) \in \mathbb{R}^{T \times d_t}. \label{eq:F_T_refined}
\end{align}

The self-attention module employs a multi-head attention mechanism followed by residual connections and layer normalization:

\begin{equation}
\text{SelfAtt}(F) = \text{LayerNorm} \left( \text{MHA}(F, F, F) + F \right)
\end{equation}

where:

\begin{itemize}
    \item \( \text{MHA}(\cdot) \) denotes the multi-head self-attention operation,
    \item \( F \) represents the input feature matrix for a given modality (\( F_V, F_A, F_T \)),
    \item \( \text{LayerNorm}(\cdot) \) ensures stable training by normalizing the updated features.
\end{itemize}

After intra-modal refinement, the features are concatenated and projected into a unified multi-modal representation:

\begin{equation}
F_{concat} = \text{Concat}(F_V^{refined}, F_A^{refined}, F_T^{refined})
\end{equation}

\begin{equation}
F_{projected} = F_{concat}W_{proj} + b_{proj} \in \mathbb{R}^{T \times d_{proj}}
\end{equation}

where \( F_{concat} \in \mathbb{R}^{T \times (d_v + d_a + d_t)} \),  
\( W_{proj} \in \mathbb{R}^{(d_v + d_a + d_t) \times d_{proj}} \),  
and \( b_{proj} \in \mathbb{R}^{d_{proj}} \) are learnable parameters.

To capture long-range dependencies and contextualized representations, we employ two BEiT-based transformer encoders~\cite{bao2021beit}. The attention mechanism is formulated as:
\begin{align}
\mathbf{d}_{t,t'} &= \tanh(\mathbf{W}_d \mathbf{h}_t + \mathbf{W}'_d \mathbf{h}_{t'} + \mathbf{b}_d), \label{eq:d_tt} \\
\alpha_{t,t'} &= \mathbf{v}_d^T \mathbf{d}_{t,t'}, \label{eq:alpha_tt} \\
a_{t,t'} &= \text{softmax}(\alpha_{t,t'}), \label{eq:a_tt} \\
\mathbf{l}_t &= \sum_{t'=1}^T a_{t,t'} \mathbf{h}_{t'}. \label{eq:l_t}
\end{align}

where:
\begin{itemize}
    \item \( \mathbf{h}_t \) and \( \mathbf{h}_{t'} \) are hidden states at positions \( t \) and \( t' \),
    \item \( \mathbf{W}_d \), \( \mathbf{W}'_d \), \( \mathbf{b}_d \), and \( \mathbf{v}_d \) are learnable parameters,
    \item \( \mathbf{l}_t \) is the refined representation at position \( t \).
\end{itemize}

The final multi-modal feature representation is obtained after passing through two consecutive BEiT-based encoders:

\begin{equation}
F_{refined} = \text{BEiT}_2(\text{BEiT}_1(F_{projected})) \in \mathbb{R}^{T \times d_{beit}}
\end{equation}

This process ensures that the final multi-modal embedding captures context-aware emotional cues, allowing for robust downstream emotion prediction.

\subsection{Polar Coordinate Emotion Prediction}

The final component of our proposed architecture is a specialized emotion prediction framework that operates in polar coordinates, aligning with the valence-arousal emotion circumplex model widely used in affective computing.

\subsubsection{Feature Pooling}
First, we apply temporal average pooling to obtain a fixed-dimensional representation:

\begin{equation}
F_{pooled} = \frac{1}{T}\sum_{t=1}^{T}F_{refined}[t] \in \mathbb{R}^{d_{beit}}
\end{equation}

\subsubsection{Emotion Predictor}
A feedforward neural network with three fully connected layers and ReLU activations processes the pooled features:
\begin{align}
h_1 &= \text{ReLU}(F_{\text{pooled}}W_1 + b_1) \in \mathbb{R}^{d_1}, \label{eq:h1} \\
h_2 &= \text{ReLU}(h_1W_2 + b_2) \in \mathbb{R}^{d_2}, \label{eq:h2} \\
[I, \theta] &= h_2W_3 + b_3 \in \mathbb{R}^{2}. \label{eq:I_theta}
\end{align}

where $W_1 \in \mathbb{R}^{d_{beit} \times d_1}$, $W_2 \in \mathbb{R}^{d_1 \times d_2}$, $W_3 \in \mathbb{R}^{d_2 \times 2}$, $b_1 \in \mathbb{R}^{d_1}$, $b_2 \in \mathbb{R}^{d_2}$, and $b_3 \in \mathbb{R}^{2}$ are learnable parameters.

\subsubsection{Polar Conversion}
The network predicts two values: intensity $I$ and angle $\theta$. These are converted to valence and arousal coordinates:

\begin{equation}
\text{valence} = I \cdot \cos(\theta)
\end{equation}

\begin{equation}
\text{arousal} = I \cdot \sin(\theta)
\end{equation}

This parametrization aligns with psychological models of emotion, where similar emotions are adjacent in the emotion circumplex. It also enforces the constraint that emotions with similar valence-arousal values have similar representations in the model's internal space.

\subsection{Training and Optimization}

The model is trained to minimize the Mean Squared Error (MSE) between predicted and ground-truth valence-arousal values:

\begin{equation}
\mathcal{L} = \frac{1}{N}\sum_{i=1}^{N}[(v_i - \hat{v}_i)^2 + (a_i - \hat{a}_i)^2]
\end{equation}

where $v_i$ and $a_i$ are the ground-truth valence and arousal values, $\hat{v}_i$ and $\hat{a}_i$ are the predicted values, and $N$ is the number of samples.

We employ the AdamW optimizer with a learning rate of $1 \times 10^{-4}$ and weight decay of $1 \times 10^{-2}$. To prevent overfitting, we apply dropout with a rate of 0.2 after each fully connected layer in the emotion predictor.

\section{Experiments and Results}
In this section, we present the experimental results to evaluate the performance of the proposed model. The analysis focuses on understanding the effectiveness of the model across CCC metric.
\begin{figure}[h]
    \centering
    \begin{tabular}{c}
        \includegraphics[width=0.9\linewidth]{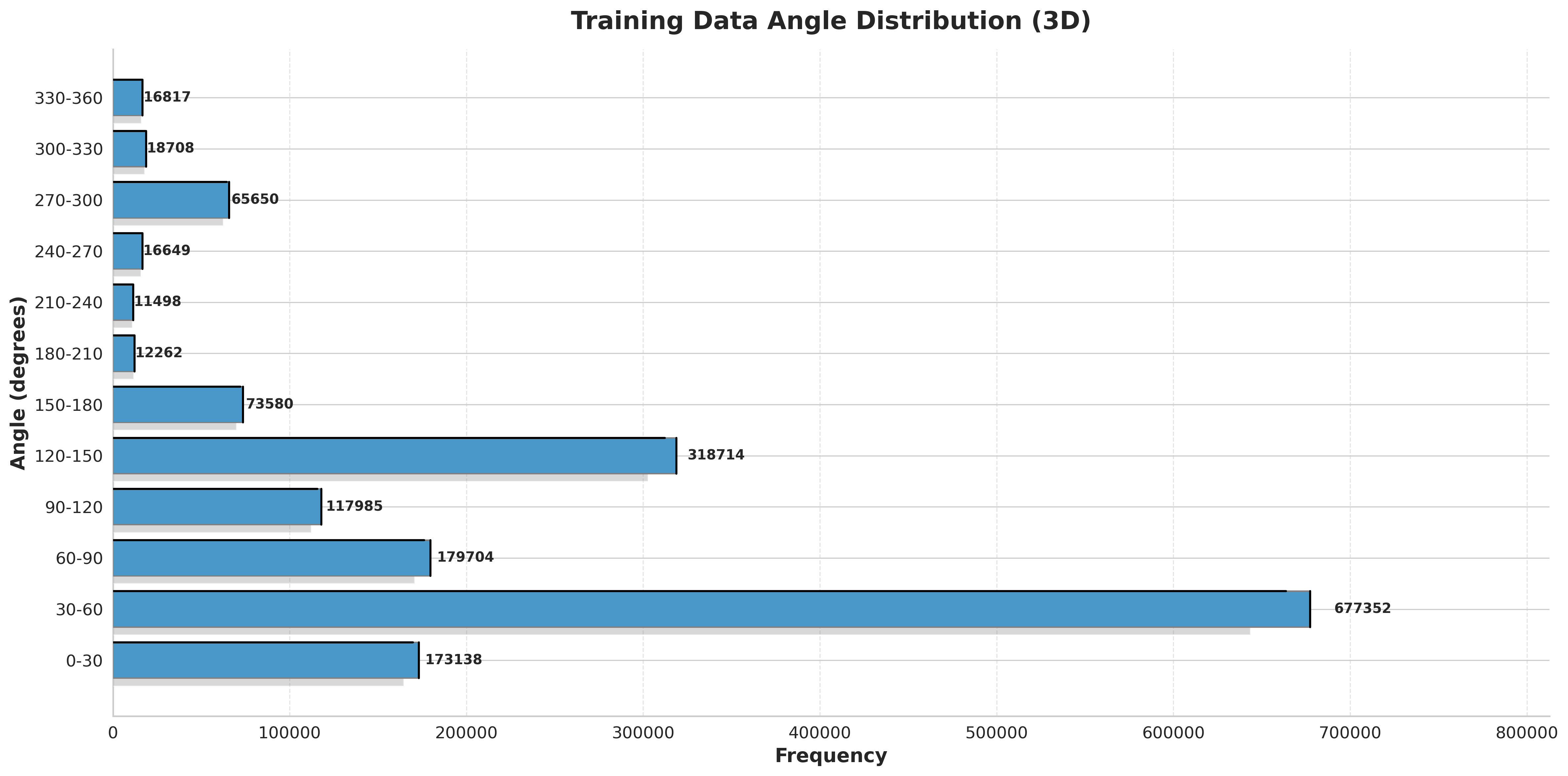} \\
        (a) Training data distribution based on angle ($\theta$) \\
        \includegraphics[width=0.9\linewidth]{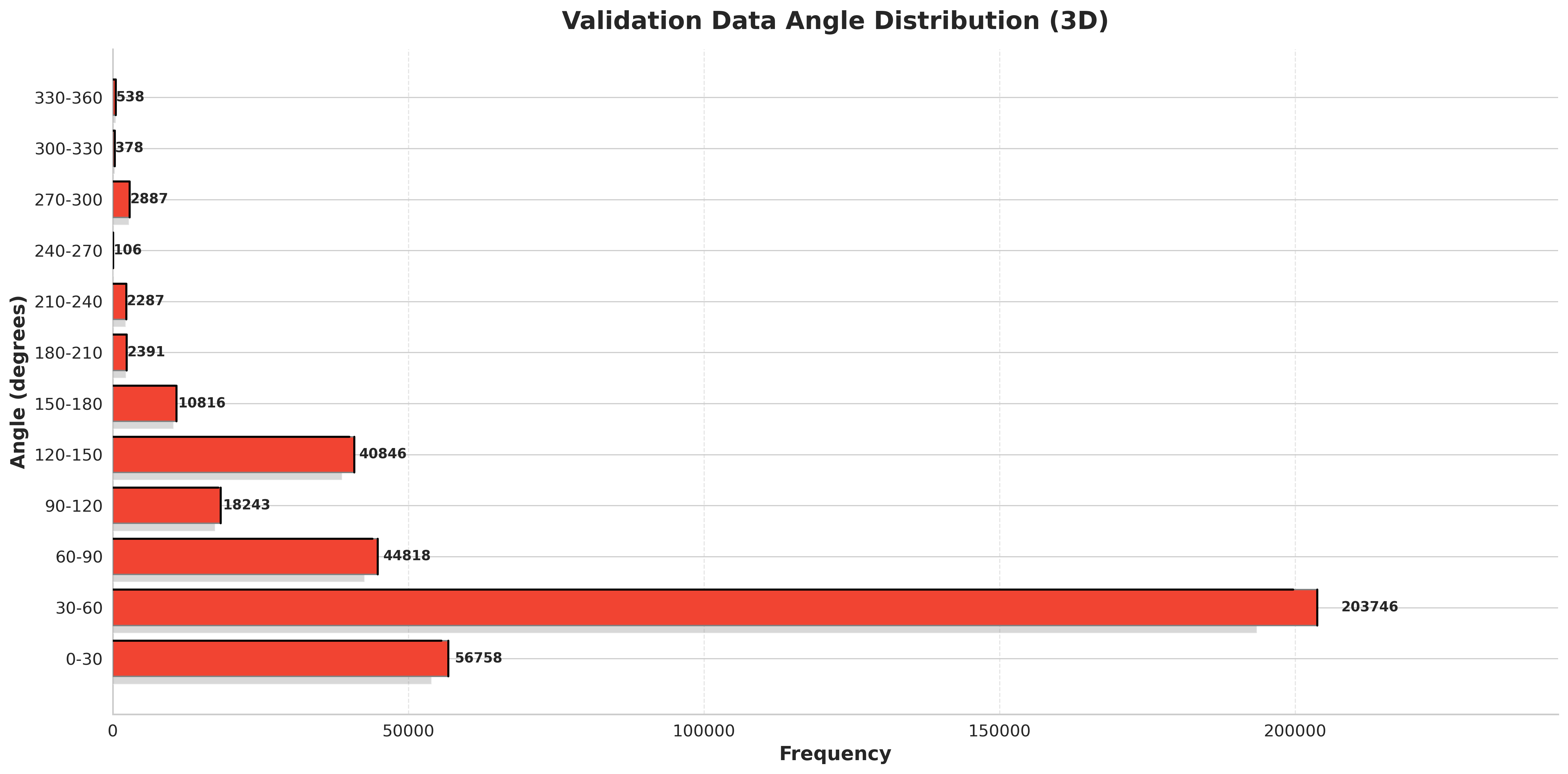} \\
        (b) Validation data distribution based on angle ($\theta$) \\
    \end{tabular}
    \caption{Example frames from the Aff-Wild2 dataset showing facial region extraction and alignment.}
    \label{fig:dataset_examples}
\end{figure}



\subsection{Dataset}
We evaluate our approach on the Aff-Wild2 dataset , the largest in-the-wild audiovisual database for valence-arousal estimation, containing 545 videos with ~2.8M frames, each annotated with continuous valence and arousal values in [-1, 1]~\cite{kollias2025advancements}. Following the ABAW competition’s train/validation/test split, we extract facial regions using a face detector, aligning them to 112$\times$112 pixels. Audio is sampled at 48kHz, and text transcriptions are generated via automatic speech recognition.

\subsection{Implementation Details}
We train the model for 200 epochs using the Adam optimizer with a learning rate of $1 \times 10^{-4}$, a batch size of 16, and a weight decay of $1 \times 10^{-3}$. The training is conducted on NVIDIA A100 GPUs.

\subsection{Evaluation Metrics}
Following the standard protocol in the ABAW competition, we use the Concordance Correlation Coefficient (CCC) as the primary evaluation metric:  

\[
\text{CCC} = \frac{2\rho\sigma_x\sigma_y}{\sigma_x^2 + \sigma_y^2 + (\mu_x - \mu_y)^2}
\]

where $\rho$ is the Pearson correlation coefficient, $\sigma_x$ and $\sigma_y$ are the standard deviations, and $\mu_x$ and $\mu_y$ are the means of the predicted and ground truth values, respectively.  

The overall performance is measured by the average CCC of valence and arousal:  

\[
\text{CCC}_{avg} = \frac{\text{CCC}_v + \text{CCC}_a}{2}
\]

\subsection{Results and Comparison}

Table \ref{tab:results_comparison1} compares MAVEN's performance against the ABAW ResNet-50 baseline.

\begin{table}[ht]
    \centering
    \small 
    \setlength{\tabcolsep}{4pt} 
    \renewcommand{\arraystretch}{0.9} 
    \caption{Results and Comparison of Different Models}
    \begin{tabular}{|c|c|c|c|}
        \hline
        \textbf{Model} & \textbf{Valence CCC} & \textbf{Arousal CCC} & \textbf{CCC Avg} \\
        \hline
        Baseline~\cite{kollias2025advancements} & 0.2400 & 0.2000 & 0.2200 \\
        \hline
        MAVEN (\textit{proposed}) & \textbf{0.3068} & \textbf{0.3054} & \textbf{0.3061} \\
        \hline
    \end{tabular}
    \label{tab:results_comparison1}
\end{table}

MAVEN achieves a CCC Avg of 0.3061, surpassing the baseline model’s 0.22, demonstrating the effectiveness of the multi-modal attention framework.

\subsection{Ablation Study}
\label{sec:ablation}

To analyze MAVEN’s design choices, we conduct ablation studies by removing individual modalities and evaluating their impact. Table~\ref{tab:ablation} presents the performance of uni-modal variants on the Aff-Wild2 dataset, measured using the Concordance Correlation Coefficient (CCC).

\begin{table}[h]
    \centering
    \caption{Results and Comparison of Different Models}
    \label{tab:ablation}
    \small
    \begin{tabular}{lccc}
        \toprule
        \textbf{Model Variant} & \textbf{Valence CCC} & \textbf{Arousal CCC} & \textbf{CCC Avg} \\
        \midrule
        Visual only & 0.1048 & 0.2459 & 0.1754 \\
        Audio only  & 0.1283 & 0.0683 & 0.0299 \\
        Text only   & 0.0019 & 0.0006 & 0.0013 \\
        \bottomrule
    \end{tabular}
\end{table}

The contributions of each modality are as follows:
\begin{itemize}
    \item \textbf{Visual Modality}: Achieves the highest average CCC (0.1754), excelling in arousal detection (0.2459) by capturing subtle facial cues like widened eyes or furrowed brows, critical for emotion recognition in conversational videos.
    \item \textbf{Audio Modality}: Yields a lower average CCC (0.0299), with better valence performance (0.1283) through prosody and pitch, but struggles in noisy in-the-wild settings, limiting its reliability.
    \item \textbf{Text Modality}: Provides minimal impact (CCC Avg: 0.0013), offering contextual cues to disambiguate emotions like sarcasm, though sparse emotional markers in transcripts constrain its contribution.
\end{itemize}

These results highlight the necessity of multi-modal integration, as MAVEN’s full model achieves a significantly higher CCC (0.3061), leveraging the synergistic strengths of all modalities.

\section{Conclusion}
This paper introduces MAVEN, a multi-modal attention architecture for valence and arousal recognition. It features a bidirectional cross-modal attention mechanism for effective information exchange, a self-attention refinement module for improved modality specificity, and a polar coordinate prediction framework for intuitive emotional representation. Achieving a SOTA CCC of \textbf{0.3061}, the model demonstrates that both the attention mechanism and polar coordinate framework significantly enhance emotion recognition by effectively utilizing complementary multi-modal cues. Future work for our proposed methodology lies in enhancing the temporal modeling capabilities of the framework to better capture the dynamics of emotion evolution.

\small
\bibliographystyle{ieeenat_fullname}
\bibliography{references}

\begin{thebibliography}{55}
\providecommand{\natexlab}[1]{#1}
\providecommand{\url}[1]{\texttt{#1}}
\expandafter\ifx\csname urlstyle\endcsname\relax
  \providecommand{\doi}[1]{doi: #1}\else
  \providecommand{\doi}{doi: \begingroup \urlstyle{rm}\Url}\fi

\bibitem[Aravindan et~al.(2023)Aravindan, Chappidi, Thumma, and Palanisamy]{aravindan2023prediction}
Abhinav~Anthiyur Aravindan, Sriram~Kalyan Chappidi, Anirudh Thumma, and Rohini Palanisamy.
\newblock Prediction of arousal and valence state from electrodermal activity using wavelet based resnet50 model.
\newblock \emph{Current Directions in Biomedical Engineering}, 9\penalty0 (1):\penalty0 555--558, 2023.

\bibitem[Bao et~al.(2021)Bao, Dong, Piao, and Wei]{bao2021beit}
Hangbo Bao, Li Dong, Songhao Piao, and Furu Wei.
\newblock Beit: Bert pre-training of image transformers.
\newblock \emph{arXiv preprint arXiv:2106.08254}, 2021.

\bibitem[Chen et~al.(2019)Chen, Zhang, Mao, Huang, Jiang, and Zhang]{chen2019accurate}
JX Chen, PW Zhang, ZJ Mao, YF Huang, DM Jiang, and YN Zhang.
\newblock Accurate eeg-based emotion recognition on combined features using deep convolutional neural networks.
\newblock \emph{Ieee Access}, 7:\penalty0 44317--44328, 2019.

\bibitem[Chen et~al.(2024)Chen, Xing, Chen, and Xu]{chen2024vesper}
Weidong Chen, Xiaofen Xing, Peihao Chen, and Xiangmin Xu.
\newblock Vesper: A compact and effective pretrained model for speech emotion recognition.
\newblock \emph{IEEE Transactions on Affective Computing}, 2024.

\bibitem[Cimtay et~al.(2020)Cimtay, Ekmekcioglu, and Caglar-Ozhan]{cimtay2020cross}
Yucel Cimtay, Erhan Ekmekcioglu, and Seyma Caglar-Ozhan.
\newblock Cross-subject multimodal emotion recognition based on hybrid fusion.
\newblock \emph{IEEE Access}, 8:\penalty0 168865--168878, 2020.

\bibitem[Dresvyanskiy et~al.(2024)Dresvyanskiy, Markitantov, Yu, Kaya, and Karpov]{dresvyanskiy2024multi}
Denis Dresvyanskiy, Maxim Markitantov, Jiawei Yu, Heysem Kaya, and Alexey Karpov.
\newblock Multi-modal arousal and valence estimation under noisy conditions.
\newblock In \emph{Proceedings of the IEEE/CVF Conference on Computer Vision and Pattern Recognition}, pages 4773--4783, 2024.

\bibitem[Fan et~al.(2024)Fan, Xie, Tao, Li, Pei, Li, and Lv]{fan2024icaps}
Cunhang Fan, Heng Xie, Jianhua Tao, Yongwei Li, Guanxiong Pei, Taihao Li, and Zhao Lv.
\newblock Icaps-reslstm: Improved capsule network and residual lstm for eeg emotion recognition.
\newblock \emph{Biomedical Signal Processing and Control}, 87:\penalty0 105422, 2024.

\bibitem[Gotsman et~al.(2021)Gotsman, Polydorou, and Edalat]{gotsman2021valence}
Tom Gotsman, Neophytos Polydorou, and Abbas Edalat.
\newblock Valence/arousal estimation of occluded faces from vr headsets.
\newblock In \emph{2021 IEEE Third International Conference on Cognitive Machine Intelligence (CogMI)}, pages 96--105. IEEE, 2021.

\bibitem[Handrich et~al.(2019)Handrich, Dinges, Saxen, Al-Hamadi, and Wachmuth]{handrich2019simultaneous}
Sebastian Handrich, Laslo Dinges, Frerk Saxen, Ayoub Al-Hamadi, and Sven Wachmuth.
\newblock Simultaneous prediction of valence/arousal and emotion categories in real-time.
\newblock In \emph{2019 IEEE International Conference on Signal and Image Processing Applications (ICSIPA)}, pages 176--180. IEEE, 2019.

\bibitem[Hsu et~al.(2021)Hsu, Bolte, Tsai, Lakhotia, Salakhutdinov, and Mohamed]{hsu2021hubert}
Wei-Ning Hsu, Benjamin Bolte, Yao-Hung~Hubert Tsai, Kushal Lakhotia, Ruslan Salakhutdinov, and Abdelrahman Mohamed.
\newblock Hubert: Self-supervised speech representation learning by masked prediction of hidden units.
\newblock \emph{IEEE/ACM transactions on audio, speech, and language processing}, 29:\penalty0 3451--3460, 2021.

\bibitem[Hwooi et~al.(2022)Hwooi, Othmani, and Sabri]{hwooi2022deep}
Stephen Khor~Wen Hwooi, Alice Othmani, and Aznul Qalid~Md Sabri.
\newblock Deep learning-based approach for continuous affect prediction from facial expression images in valence-arousal space.
\newblock \emph{IEEE Access}, 10:\penalty0 96053--96065, 2022.

\bibitem[Khan et~al.(2024)Khan, Gueaieb, El~Saddik, and Kwon]{khan2024mser}
Mustaqeem Khan, Wail Gueaieb, Abdulmotaleb El~Saddik, and Soonil Kwon.
\newblock Mser: Multimodal speech emotion recognition using cross-attention with deep fusion.
\newblock \emph{Expert Systems with Applications}, 245:\penalty0 122946, 2024.

\bibitem[Kollias(2022)]{kollias2022abaw}
Dimitrios Kollias.
\newblock Abaw: Valence-arousal estimation, expression recognition, action unit detection \& multi-task learning challenges.
\newblock In \emph{Proceedings of the IEEE/CVF Conference on Computer Vision and Pattern Recognition}, pages 2328--2336, 2022.

\bibitem[Kollias(2023)]{kollias2023multi}
Dimitrios Kollias.
\newblock Multi-label compound expression recognition: C-expr database \& network.
\newblock In \emph{Proceedings of the IEEE/CVF Conference on Computer Vision and Pattern Recognition}, pages 5589--5598, 2023.

\bibitem[Kollias and Zafeiriou(2018)]{kollias2018multi}
Dimitrios Kollias and Stefanos Zafeiriou.
\newblock A multi-task learning \& generation framework: Valence-arousal, action units \& primary expressions.
\newblock \emph{arXiv preprint arXiv:1811.07771}, 2018.

\bibitem[Kollias and Zafeiriou(2019)]{kollias2019expression}
Dimitrios Kollias and Stefanos Zafeiriou.
\newblock Expression, affect, action unit recognition: Aff-wild2, multi-task learning and arcface.
\newblock \emph{arXiv preprint arXiv:1910.04855}, 2019.

\bibitem[Kollias and Zafeiriou(2020)]{kollias2020exploiting}
Dimitrios Kollias and Stefanos Zafeiriou.
\newblock Exploiting multi-cnn features in cnn-rnn based dimensional emotion recognition on the omg in-the-wild dataset.
\newblock \emph{IEEE Transactions on Affective Computing}, 12\penalty0 (3):\penalty0 595--606, 2020.

\bibitem[Kollias and Zafeiriou(2021{\natexlab{a}})]{kollias2021affect}
Dimitrios Kollias and Stefanos Zafeiriou.
\newblock Affect analysis in-the-wild: Valence-arousal, expressions, action units and a unified framework.
\newblock \emph{arXiv preprint arXiv:2103.15792}, 2021{\natexlab{a}}.

\bibitem[Kollias and Zafeiriou(2021{\natexlab{b}})]{kollias2021analysing}
Dimitrios Kollias and Stefanos Zafeiriou.
\newblock Analysing affective behavior in the second abaw2 competition.
\newblock In \emph{Proceedings of the IEEE/CVF International Conference on Computer Vision}, pages 3652--3660, 2021{\natexlab{b}}.

\bibitem[Kollias et~al.(2019{\natexlab{a}})Kollias, Sharmanska, and Zafeiriou]{kollias2019face}
Dimitrios Kollias, Viktoriia Sharmanska, and Stefanos Zafeiriou.
\newblock Face behavior a la carte: Expressions, affect and action units in a single network.
\newblock \emph{arXiv preprint arXiv:1910.11111}, 2019{\natexlab{a}}.

\bibitem[Kollias et~al.(2019{\natexlab{b}})Kollias, Tzirakis, Nicolaou, Papaioannou, Zhao, Schuller, Kotsia, and Zafeiriou]{kollias2019deep}
Dimitrios Kollias, Panagiotis Tzirakis, Mihalis~A Nicolaou, Athanasios Papaioannou, Guoying Zhao, Bj{\"o}rn Schuller, Irene Kotsia, and Stefanos Zafeiriou.
\newblock Deep affect prediction in-the-wild: Aff-wild database and challenge, deep architectures, and beyond.
\newblock \emph{International Journal of Computer Vision}, 127\penalty0 (6):\penalty0 907--929, 2019{\natexlab{b}}.

\bibitem[Kollias et~al.(2020)Kollias, Schulc, Hajiyev, and Zafeiriou]{kollias2020analysing}
Dimitrios Kollias, Attila Schulc, Elnar Hajiyev, and Stefanos Zafeiriou.
\newblock Analysing affective behavior in the first abaw 2020 competition.
\newblock In \emph{2020 15th IEEE International Conference on Automatic Face and Gesture Recognition (FG 2020)}, pages 637--643. IEEE, 2020.

\bibitem[Kollias et~al.(2021)Kollias, Sharmanska, and Zafeiriou]{kollias2021distribution}
Dimitrios Kollias, Viktoriia Sharmanska, and Stefanos Zafeiriou.
\newblock Distribution matching for heterogeneous multi-task learning: a large-scale face study.
\newblock \emph{arXiv preprint arXiv:2105.03790}, 2021.

\bibitem[Kollias et~al.(2023)Kollias, Tzirakis, Baird, Cowen, and Zafeiriou]{kollias2023abaw}
Dimitrios Kollias, Panagiotis Tzirakis, Alice Baird, Alan Cowen, and Stefanos Zafeiriou.
\newblock Abaw: Valence-arousal estimation, expression recognition, action unit detection \& emotional reaction intensity estimation challenges.
\newblock In \emph{Proceedings of the IEEE/CVF Conference on Computer Vision and Pattern Recognition}, pages 5889--5898, 2023.

\bibitem[Kollias et~al.(2024{\natexlab{a}})Kollias, Sharmanska, and Zafeiriou]{kollias2024distribution}
Dimitrios Kollias, Viktoriia Sharmanska, and Stefanos Zafeiriou.
\newblock Distribution matching for multi-task learning of classification tasks: a large-scale study on faces \& beyond.
\newblock In \emph{Proceedings of the AAAI Conference on Artificial Intelligence}, pages 2813--2821, 2024{\natexlab{a}}.

\bibitem[Kollias et~al.(2024{\natexlab{b}})Kollias, Tzirakis, Cowen, Zafeiriou, Kotsia, Baird, Gagne, Shao, and Hu]{kollias20246th}
Dimitrios Kollias, Panagiotis Tzirakis, Alan Cowen, Stefanos Zafeiriou, Irene Kotsia, Alice Baird, Chris Gagne, Chunchang Shao, and Guanyu Hu.
\newblock The 6th affective behavior analysis in-the-wild (abaw) competition.
\newblock In \emph{Proceedings of the IEEE/CVF Conference on Computer Vision and Pattern Recognition}, pages 4587--4598, 2024{\natexlab{b}}.

\bibitem[Kollias et~al.(2024{\natexlab{c}})Kollias, Zafeiriou, Kotsia, Dhall, Ghosh, Shao, and Hu]{kollias20247th}
Dimitrios Kollias, Stefanos Zafeiriou, Irene Kotsia, Abhinav Dhall, Shreya Ghosh, Chunchang Shao, and Guanyu Hu.
\newblock 7th abaw competition: Multi-task learning and compound expression recognition.
\newblock \emph{arXiv preprint arXiv:2407.03835}, 2024{\natexlab{c}}.

\bibitem[Kollias et~al.(2025)Kollias, Tzirakis, Cowen, Zafeiriou, Kotsia, Granger, Pedersoli, Bacon, Baird, Gagne, et~al.]{kollias2025advancements}
Dimitrios Kollias, Panagiotis Tzirakis, AS Cowen, S Zafeiriou, I Kotsia, Eric Granger, Marco Pedersoli, SL Bacon, Alice Baird, C Gagne, et~al.
\newblock Advancements in affective and behavior analysis: The 8th abaw workshop and competition.
\newblock 2025.

\bibitem[Liu et~al.(2023)Liu, Sun, Jiang, Zhang, Deng, Huang, Meng, Liu, and Liu]{liu2023evaef}
Xiaolong Liu, Lei Sun, Wenqiang Jiang, Fengyuan Zhang, Yuanyuan Deng, Zhaopei Huang, Liyu Meng, Yuchen Liu, and Chuanhe Liu.
\newblock Evaef: Ensemble valence-arousal estimation framework in the wild.
\newblock In \emph{Proceedings of the IEEE/CVF Conference on Computer Vision and Pattern Recognition}, pages 5863--5871, 2023.

\bibitem[Liu et~al.(2019)Liu, Ott, Goyal, Du, Joshi, Chen, Levy, Lewis, Zettlemoyer, and Stoyanov]{liu2019roberta}
Yinhan Liu, Myle Ott, Naman Goyal, Jingfei Du, Mandar Joshi, Danqi Chen, Omer Levy, Mike Lewis, Luke Zettlemoyer, and Veselin Stoyanov.
\newblock Roberta: A robustly optimized bert pretraining approach.
\newblock \emph{arXiv preprint arXiv:1907.11692}, 2019.

\bibitem[Liu et~al.(2021)Liu, Lin, Cao, Hu, Wei, Zhang, Lin, and Guo]{liu2021swin}
Ze Liu, Yutong Lin, Yue Cao, Han Hu, Yixuan Wei, Zheng Zhang, Stephen Lin, and Baining Guo.
\newblock Swin transformer: Hierarchical vision transformer using shifted windows.
\newblock In \emph{Proceedings of the IEEE/CVF international conference on computer vision}, pages 10012--10022, 2021.

\bibitem[Meng et~al.(2022{\natexlab{a}})Meng, Liu, Liu, Huang, Cheng, Wang, Liu, and Jin]{meng2022multi}
Liyu Meng, Yuchen Liu, Xiaolong Liu, Zhaopei Huang, Yuan Cheng, Meng Wang, Chuanhe Liu, and Qin Jin.
\newblock Multi-modal emotion estimation for in-the-wild videos.
\newblock \emph{arXiv preprint arXiv:2203.13032}, 2022{\natexlab{a}}.

\bibitem[Meng et~al.(2022{\natexlab{b}})Meng, Liu, Liu, Huang, Jiang, Zhang, Liu, and Jin]{meng2022valence}
Liyu Meng, Yuchen Liu, Xiaolong Liu, Zhaopei Huang, Wenqiang Jiang, Tenggan Zhang, Chuanhe Liu, and Qin Jin.
\newblock Valence and arousal estimation based on multimodal temporal-aware features for videos in the wild.
\newblock In \emph{Proceedings of the IEEE/CVF Conference on Computer Vision and Pattern Recognition}, pages 2345--2352, 2022{\natexlab{b}}.

\bibitem[Mitenkova et~al.(2019)Mitenkova, Kossaifi, Panagakis, and Pantic]{mitenkova2019valence}
Anna Mitenkova, Jean Kossaifi, Yannis Panagakis, and Maja Pantic.
\newblock Valence and arousal estimation in-the-wild with tensor methods.
\newblock In \emph{2019 14th IEEE International Conference on Automatic Face \& Gesture Recognition (FG 2019)}, pages 1--7. IEEE, 2019.

\bibitem[Mou et~al.(2015)Mou, Celiktutan, and Gunes]{mou2015group}
Wenxuan Mou, Oya Celiktutan, and Hatice Gunes.
\newblock Group-level arousal and valence recognition in static images: Face, body and context.
\newblock In \emph{2015 11th IEEE International Conference and Workshops on Automatic Face and Gesture Recognition (FG)}, pages 1--6. IEEE, 2015.

\bibitem[Pinto et~al.(2020)Pinto, Gon{\c{c}}alves, Pinto, Sanhudo, Fonseca, Gon{\c{c}}alves, Carvalho, and Cardoso]{pinto2020audiovisual}
Jo{\~a}o~Ribeiro Pinto, Tiago Gon{\c{c}}alves, Carolina Pinto, Lu{\'\i}s Sanhudo, Joaquim Fonseca, Filipe Gon{\c{c}}alves, Pedro Carvalho, and Jaime~S Cardoso.
\newblock Audiovisual classification of group emotion valence using activity recognition networks.
\newblock In \emph{2020 IEEE 4th International Conference on Image Processing, Applications and Systems (IPAS)}, pages 114--119. IEEE, 2020.

\bibitem[Praveen et~al.(2022)Praveen, de~Melo, Ullah, Aslam, Zeeshan, Denorme, Pedersoli, Koerich, Bacon, Cardinal, et~al.]{praveen2022joint}
R~Gnana Praveen, Wheidima~Carneiro de Melo, Nasib Ullah, Haseeb Aslam, Osama Zeeshan, Th{\'e}o Denorme, Marco Pedersoli, Alessandro~L Koerich, Simon Bacon, Patrick Cardinal, et~al.
\newblock A joint cross-attention model for audio-visual fusion in dimensional emotion recognition.
\newblock In \emph{Proceedings of the IEEE/CVF conference on computer vision and pattern recognition}, pages 2486--2495, 2022.

\bibitem[Russell(1980)]{russell1980circumplex}
James~A Russell.
\newblock A circumplex model of affect.
\newblock \emph{Journal of personality and social psychology}, 39\penalty0 (6):\penalty0 1161, 1980.

\bibitem[Savchenko(2022)]{savchenko2022frame}
Andrey~V Savchenko.
\newblock Frame-level prediction of facial expressions, valence, arousal and action units for mobile devices.
\newblock \emph{arXiv preprint arXiv:2203.13436}, 2022.

\bibitem[Savchenko(2023)]{savchenko2023emotieffnets}
Andrey~V Savchenko.
\newblock Emotieffnets for facial processing in video-based valence-arousal prediction, expression classification and action unit detection.
\newblock In \emph{Proceedings of the IEEE/CVF Conference on Computer Vision and Pattern Recognition}, pages 5716--5724, 2023.

\bibitem[Savchenko(2024)]{savchenko2024hsemotion}
Andrey~V Savchenko.
\newblock Hsemotion team at the 6th abaw competition: Facial expressions, valence-arousal and emotion intensity prediction.
\newblock \emph{arXiv preprint arXiv:2403.11590}, 2024.

\bibitem[Savchenko(2025)]{savchenko2025hsemotion}
Andrey~V Savchenko.
\newblock Hsemotion team at abaw-8 competition: Audiovisual ambivalence/hesitancy, emotional mimicry intensity and facial expression recognition.
\newblock \emph{arXiv preprint arXiv:2503.10399}, 2025.

\bibitem[Savran et~al.(2013)Savran, Gur, and Verma]{savran2013automatic}
Arman Savran, Ruben Gur, and Ragini Verma.
\newblock Automatic detection of emotion valence on faces using consumer depth cameras.
\newblock In \emph{Proceedings of the IEEE International Conference on Computer Vision Workshops}, pages 75--82, 2013.

\bibitem[Song et~al.(2018)Song, Liu, Wang, Hang, and Huang]{song2018spatiotemporal}
Huihui Song, Qingshan Liu, Guojie Wang, Renlong Hang, and Bo Huang.
\newblock Spatiotemporal satellite image fusion using deep convolutional neural networks.
\newblock \emph{IEEE Journal of Selected Topics in Applied Earth Observations and Remote Sensing}, 11\penalty0 (3):\penalty0 821--829, 2018.

\bibitem[Theagarajan et~al.(2018)Theagarajan, Bhanu, and Cruz]{theagarajan2018deepdriver}
Rajkumar Theagarajan, Bir Bhanu, and Albert Cruz.
\newblock Deepdriver: Automated system for measuring valence and arousal in car driver videos.
\newblock In \emph{2018 24th International Conference on Pattern Recognition (ICPR)}, pages 2546--2551. IEEE, 2018.

\bibitem[Verma and Tiwary(2017)]{verma2017affect}
Gyanendra~K Verma and Uma~Shanker Tiwary.
\newblock Affect representation and recognition in 3d continuous valence--arousal--dominance space.
\newblock \emph{Multimedia Tools and Applications}, 76:\penalty0 2159--2183, 2017.

\bibitem[Wei et~al.(2024)Wei, Hu, Yang, Luu, and Dong]{wei2024learning}
Jie Wei, Guanyu Hu, Xinyu Yang, Anh~Tuan Luu, and Yizhuo Dong.
\newblock Learning facial expression and body gesture visual information for video emotion recognition.
\newblock \emph{Expert Systems with Applications}, 237:\penalty0 121419, 2024.

\bibitem[Xie et~al.(2021)Xie, Li, Lo, Shuai, Cheng, et~al.]{xie2021technical}
Hong-Xia Xie, I Li, Ling Lo, Hong-Han Shuai, Wen-Huang Cheng, et~al.
\newblock Technical report for valence-arousal estimation in abaw2 challenge.
\newblock \emph{arXiv preprint arXiv:2107.03891}, 2021.

\bibitem[Yanagimoto and Sugimoto(2016)]{yanagimoto2016recognition}
Miku Yanagimoto and Chika Sugimoto.
\newblock Recognition of persisting emotional valence from eeg using convolutional neural networks.
\newblock In \emph{2016 IEEE 9th International Workshop on Computational Intelligence and Applications (IWCIA)}, pages 27--32. IEEE, 2016.

\bibitem[Yang and Lee(2019)]{yang2019attribute}
Hao-Chun Yang and Chi-Chun Lee.
\newblock An attribute-invariant variational learning for emotion recognition using physiology.
\newblock In \emph{ICASSP 2019-2019 IEEE international conference on acoustics, speech and signal processing (ICASSP)}, pages 1184--1188. IEEE, 2019.

\bibitem[Yu et~al.(2024)Yu, Zhao, Wang, Wei, Zhang, Cai, Xie, Zhu, Zhu, Yang, et~al.]{yu2024improving}
Jun Yu, Gongpeng Zhao, Yongqi Wang, Zhihong Wei, Zerui Zhang, Zhongpeng Cai, Guochen Xie, Jichao Zhu, Wangyuan Zhu, Shuoping Yang, et~al.
\newblock Improving valence-arousal estimation with spatiotemporal relationship learning and multimodal fusion.
\newblock In \emph{Proceedings of the IEEE/CVF Conference on Computer Vision and Pattern Recognition}, pages 7878--7885, 2024.

\bibitem[Zafeiriou et~al.(2017)Zafeiriou, Kollias, Nicolaou, Papaioannou, Zhao, and Kotsia]{zafeiriou2017aff}
Stefanos Zafeiriou, Dimitrios Kollias, Mihalis~A Nicolaou, Athanasios Papaioannou, Guoying Zhao, and Irene Kotsia.
\newblock Aff-wild: valence and arousal'in-the-wild'challenge.
\newblock In \emph{Proceedings of the IEEE conference on computer vision and pattern recognition workshops}, pages 34--41, 2017.

\bibitem[Zhang et~al.(2024)Zhang, Yang, Chen, Zhang, Leng, and Zhao]{zhang2024deep}
Shiqing Zhang, Yijiao Yang, Chen Chen, Xingnan Zhang, Qingming Leng, and Xiaoming Zhao.
\newblock Deep learning-based multimodal emotion recognition from audio, visual, and text modalities: A systematic review of recent advancements and future prospects.
\newblock \emph{Expert Systems with Applications}, 237:\penalty0 121692, 2024.

\bibitem[Zhang et~al.(2023)Zhang, An, Cui, Xu, Dong, Jiang, Shi, Liu, Sun, and Wang]{zhang2023abaw5}
Ziyang Zhang, Liuwei An, Zishun Cui, Ao Xu, Tengteng Dong, Yueqi Jiang, Jingyi Shi, Xin Liu, Xiao Sun, and Meng Wang.
\newblock Abaw5 challenge: A facial affect recognition approach utilizing transformer encoder and audiovisual fusion.
\newblock In \emph{Proceedings of the IEEE/CVF Conference on Computer Vision and Pattern Recognition}, pages 5725--5734, 2023.

\bibitem[Zhou et~al.(2025)Zhou, Ling, and Cai]{zhou2025emotion}
Weiwei Zhou, Chenkun Ling, and Zefeng Cai.
\newblock Emotion recognition with clip and sequential learning.
\newblock \emph{arXiv preprint arXiv:2503.09929}, 2025.

\end{thebibliography}
\nocite{*}


\end{document}